\documentclass{article} 
\usepackage{nips11submit_e,times}

\usepackage{tikz,amsmath,amssymb,booktabs}
\usepackage{pgfplots}

\usepackage[square,numbers]{natbib}

\usepackage{microtype}

\usepackage[ruled,vlined,linesnumbered]{algorithm2e}

\usetikzlibrary{arrows,shapes,plotmarks}

\tikzset{>=stealth'} 
\tikzstyle{graphnode} = 
   [circle,draw=black,minimum size=22pt,text centered,text
     width=22pt,inner sep=0pt] 
\tikzstyle{obs} = [graphnode,fill=black,text=white]
\tikzstyle{var} = [graphnode,fill=white,text=black]
\tikzstyle{fac} = [rectangle,draw=black,fill=black!25,minimum size=5pt]
\tikzstyle{edge}= [draw=white,double=black,thick,-]

\newcommand{\g}{\,|\,}
\newcommand{\de}{\partial}
\renewcommand{\d}{\:\mathrm{d}}

\renewcommand{\H}{\mathcal{H}}
\newcommand{\K}{\mathcal{K}}
\renewcommand{\Re}{\mathbb{R}}
\newcommand{\N}{\mathcal{N}}
\newcommand{\Trans}{^{\intercal}}

\newcommand{\eval}{\text{eval}}

\newcommand{\SE}{\text{SE}}

\renewcommand{\vec}{\boldsymbol}

\renewcommand{\O}{\mathcal{O}}
\newcommand{\GP}{\mathcal{GP}}
\newcommand{\Id}{\vec{I}}
\newcommand{\tr}{\operatorname{tr}}

\newcommand{\sA}{\boldsymbol{\mathsf{A}}}
\newcommand{\sB}{\boldsymbol{\mathsf{B}}}
\newcommand{\sC}{\boldsymbol{\mathsf{C}}}
\newcommand{\sD}{\boldsymbol{\mathsf{D}}}

\DeclareSymbolFont{stmry}{U}{stmry}{m}{n}
\DeclareMathSymbol\leftarrowtriangle\mathrel{stmry}{"5E}
\DeclareMathSymbol\rightarrowtriangle\mathrel{stmry}{"5F}

\renewcommand{\to}{\rightarrowtriangle}

\title{Optimal Reinforcement Learning\\for Gaussian Systems}

\author{
Philipp Hennig\\
Max Planck Institute for Intelligent Systems\\
Department of Empirical Inference\\
Spemannstra\ss e 38, 72070 T\"ubingen, Germany \\
\texttt{phennig@tuebingen.mpg.de} \\
}

\nipsfinalcopy % Uncomment for camera-ready version

\begin{document}

\maketitle

\begin{abstract}
  The exploration-exploitation trade-off is among the central
  challenges of reinforcement learning. The optimal Bayesian solution
  is intractable in general. This paper studies to what extent
  analytic statements about optimal learning are possible if all
  beliefs are Gaussian processes. A first order approximation of
  learning of both loss and dynamics, for nonlinear, time-varying
  systems in continuous time and space, subject to a relatively weak
  restriction on the dynamics, is described by an infinite-dimensional
  partial differential equation. An approximate finite-dimensional
  projection gives an impression for how this result may be helpful.
\end{abstract}

\section{Introduction -- Optimal Reinforcement Learning}
\label{sec:intr-optim-reinf}

Reinforcement learning is about doing two things at once:
\emph{Optimizing} a function while \emph{learning} about it. These two
objectives must be balanced: Ignorance precludes efficient
optimization; time spent hunting after irrelevant knowledge incurs
unnecessary loss. This dilemma is famously known as the
\emph{exploration exploitation trade-off}. Classic reinforcement
learning often considers time cheap; the trade-off then plays a
subordinate role to the desire for learning a ``correct'' model or
policy. Many classic reinforcement learning algorithms thus rely on
ad-hoc methods to control exploration, such as ``$\epsilon$-greedy''
\citep{SuttonBarto}, or ``Thompson sampling''
\citep{Thompson}. However, at least since a thesis by Duff
\citep{duff2002optimal} it has been known that Bayesian inference
allows optimal balance between exploration and exploitation. It
requires integration over every possible future trajectory under the
current belief about the system's dynamics, all possible new data
acquired along those trajectories, and their effect on decisions taken
along the way. This amounts to optimization and integration over a
tree, of exponential cost in the size of the state space
\citep{poupart2006analytic}. The situation is particularly dire for
continuous space-times, where both depth and branching factor of the
``tree'' are uncountably infinite. Several authors have proposed
approximating this lookahead through samples
\citep{Dearden99modelbased,strens2000bayesian,wang2005bayesian,asmuth2009bayesian},
or ad-hoc estimators that can be shown to be in some sense close to
the Bayes-optimal policy \citep{kolter2009near}.

% When the state space is finite and discrete, bound-based reasoning is
% possible, which can guarantee that, at least, the resulting algorithms
% always over-explore, never under-explore
% \citep{Lai19854,auer2002finite,kearns2002near,brafman2003r,strehl2008analysis}. But
% bound-based algorithms can not be extended to continuous spaces
% without making further assumptions \citep{kleinberg2008multi}, and
% doing so invalidates the strongest argument in their favour --- that
% they are free of assumptions.
%
In a parallel development, recent work by Todorov
\citep{todorov2007linearly}, Kappen \citep{Kappen_PathIntegrals} and
others introduced an idea to reinforcement learning long commonplace
in other areas of machine learning: Structural assumptions, while
restrictive, can greatly simplify inference problems. In particular, a
recent paper by Simpkins et al. \citep{simpkins2008optimal} showed
that it is actually possible to solve the exploration exploitation
trade-off \emph{locally}, by constructing a \emph{linear}
approximation using a Kalman filter. Simpkins and colleagues further
assumed to know the loss function, and the dynamics up to Brownian
drift. Here, I use their work as inspiration for a study of general
optimal reinforcement learning of dynamics \emph{and} loss functions
of an unknown, \emph{nonlinear, time-varying} system (note that most
reinforcement learning algorithms are restricted to time-invariant
systems). The core assumption is that all uncertain variables are
known up to Gaussian process uncertainty. The main result is a
first-order description of optimal reinforcement learning in form of
infinite-dimensional differential statements. This kind of description
opens up new approaches to reinforcement learning. As an only initial
example of such treatments, Section \ref{sec:numer-solv-hamilt}
presents an approximate Ansatz that affords an explicit reinforcement
learning algorithm; tested in some simple but instructive experiments
(Section \ref{sec:experiments}).

An intuitive description of the paper's results is this: From prior
and corresponding choice of learning machinery (Section
\ref{sec:class-learn-probl}), we construct statements about the
\emph{dynamics of the learning process} (Section
\ref{sec:optim-contr-learn}). The learning machine itself provides a
probabilistic description of the dynamics of the physical
system. Combining both dynamics yields a \emph{joint} system, which we
aim to control optimally. Doing so amounts to simultaneously
controlling exploration (controlling the learning system) and
exploitation (controlling the physical system).

Because large parts of the analysis rely on concepts from optimal
control theory, this paper will use notation from that field. Readers
more familiar with the reinforcement learning literature may wish to
mentally replace coordinates $x$ with states $s$, controls $u$ with
actions $a$, dynamics with transitions $p(s'\g s,a)$ and utilities $q$
with \emph{losses} (negative rewards) $-r$. The latter is potentially
confusing, so note that optimal control in this paper will attempt to
\emph{minimize} values, rather than to maximize them, as usual in
reinforcement learning (these two descriptions are, of course,
equivalent).

\section{A Class of Learning Problems}
\label{sec:class-learn-probl}

We consider the task of optimally controlling an uncertain system
whose states $s\equiv (x,t)\in\K\equiv\Re^D\times\Re$ lie in a $D+1$
dimensional Euclidean phase space-time: A cost $Q$ (cumulated loss) is
acquired at $(x,t)$ with \emph{rate} $\mathrm{d} Q/\mathrm{d} t =
q(x,t)$, and the first inference problem is to learn this analytic
function $q$. A second, independent learning problem concerns the
dynamics of the system. We assume the dynamics separate into
\emph{free} and \emph{controlled} terms affine to the control:
\begin{equation}
  \label{eq:1}
  \d x(t) = [f(x,t) + g(x,t) u(x,t)] \d t % + \Omega_x \d \omega
\end{equation}
where $u(x,t)$ is the control function we seek to optimize, and $f,g$
are analytic functions. To simplify our analysis, we will assume that
\emph{either} $f$ \emph{or} $g$ are known, while the other may be
uncertain (or, alternatively, that it is possible to obtain
independent samples from both functions). See Section
\ref{sec:relax-some-assumpt} for a note on how this assumption may be
relaxed. W.l.o.g., let $f$ be uncertain and $g$ known. %
% $\Omega_x$ is a noise scaling matrix, and $\d \omega$ is the Wiener
% measure (i.e.\ we assume the system is stochastic, with coloured
% Gaussian noise of known structure).
Information about both $q(x,t)$ and $f(x,t)=[f_1,\dots,f_D]$ is
acquired stochastically: A Poisson process of constant rate $\lambda$
produces mutually independent samples
\begin{equation}
  \label{eq:2}
  y_q(x,t) = q(x,t) + \epsilon_q \text{~and~} y_{fd}(x,t) = f_d(x,t)
  + \epsilon_{fd} \text{~where~}  \epsilon_q \sim \N(0,\sigma_q ^2); \epsilon_{fd}
  \sim\N(0,\sigma_{fd} ^2).
\end{equation}
The noise levels $\sigma_q$ and $\sigma_f$ are presumed known. Let our
initial beliefs about $q$ and $f$ be given by Gaussian processes
$\GP_{k_q}(q;\mu_q,\Sigma_q);$ and independent Gaussian processes
$\prod_d ^D \GP_{k_{fd}}(f_d;\mu_{fd},\Sigma_{fd})$, respectively,
with kernels $k_r,k_{f1},\dots,k_{fD}$ over $\K$, and mean /
covariance functions $\mu$ / $\Sigma$. In other words, samples over
the belief can be drawn using an infinite vector $\Omega$ of
i.i.d. Gaussian variables, as
\begin{equation}
  \label{eq:4}
  \tilde{f}_d([x,t]) = \mu_{fd}([x,t]) + \int\Sigma_{fd}
  ^{1/2}([x,t],[x',t']) \Omega(x',t') \d x' \d t = \mu_{fd}([x,t]) + (\Sigma_{fd}
  ^{1/2} \Omega)([x,t])
\end{equation}
the second equation demonstrates a compact notation for inner products
that will be used throughout. It is important to note that $f,q$ are
unknown, but deterministic. At any point during learning, we can use
the same samples $\Omega$ to describe uncertainty, while
$\mu,\Sigma$ change during the learning process.

To ensure continuous trajectories, we also need to regularize the
control. Following control custom, we introduce a quadratic
control cost $\rho(u) = \frac{1}{2} u\Trans R^{-1} u$ with
control cost scaling matrix $R$. Its units $[R]=[x/t] / [Q/x]$ relate
the cost of changing location to the utility gained by doing so.

The overall task is to find the optimal discounted horizon
\emph{value}
\begin{equation}
  \label{eq:3}
  v(x,t) = \min_{u} \int_t ^\infty e^{-(\tau-t)/\gamma}
  \left[q[\chi[\tau,u(\chi,\tau)],\tau] 
    + \frac{1}{2}u(\chi,\tau)\Trans R^{-1}u(\chi,\tau)\right] \d \tau
\end{equation}
where $\chi(\tau,u)$ is the trajectory generated by the dynamics defined
in Equation \eqref{eq:1}, using the control law (policy) $u(x,t)$. The
exponential definition of the discount $\gamma>0$ gives the unit of
time to $\gamma$.

Before beginning the analysis, consider the relative generality of
this definition: We allow for a continuous phase space. Both loss
\emph{and} dynamics may be uncertain, of rather general nonlinear
form, and may change over time. The specific choice of a Poisson
process for the generation of samples is somewhat ad-hoc, but
\emph{some} measure is required to quantify the flow of information
through time. The Poisson process is in some sense the simplest such
measure, assigning uniform probability density. An alternative is to
assume that datapoints are acquired at regular intervals of width
$\lambda$. This results in a quite similar model but, since the
system's dynamics still proceed in continuous time, can complicate
notation. A downside is that we had to restrict the form of the
dynamics. However, Eq. \eqref{eq:1} still covers numerous physical
systems studied in control, for example many mechanical systems, from
classics like cart-and-pole to realistic models for helicopters
\citep{Fantoni}.

\section{Optimal Control for the Learning Process}
\label{sec:optim-contr-learn}

The optimal solution to the exploration exploitation trade-off is
formed by the \emph{dual control} \citep{FeldbaumDualControl} of a
joint representation of the physical system and the beliefs over
it. In reinforcement learning, this idea is known as a
belief-augmented POMDP \citep{duff2002optimal,poupart2006analytic},
but is not usually construed as a control problem. This section
constructs the Hamilton-Jacobi-Bellman (HJB) equation of the joint
control problem for the system described in Sec.\
\ref{sec:class-learn-probl}, and analytically solves the equation for
the optimal control. This necessitates a description of the learning
algorithm's dynamics:

%\subsection{Dynamics of Gaussian Process Inference}
%\label{sec:dynam-gauss-proc}
%
At time $t=\tau$, let the system be at phase space-time $s_\tau =
(x(\tau),\tau)$ and have the Gaussian process belief
$\GP(q;\mu_\tau(s),\Sigma_\tau(s,s'))$ over the function $q$ (all
derivations in this section will focus on $q$, and we will drop the
sub-script $q$ from many quantities for readability. The forms for
$f$, or $g$, are entirely analogous, with independent Gaussian
processes for each dimension $d=1,\dots,D$). This belief stems from a
finite number $N$ of samples $\vec{y}_{0}=[y_{1},\dots,y_{N}]\Trans
\in \Re^{N}$ collected at space-times
$\vec{S}_{0}=[(x_{1},t_1),\dots,(x_{N},t_N)]\Trans\equiv[s_1,\dots,s_N]\Trans\in
\K^{N}$ (note that $t_{1}$ to $t_{N}$ need not be equally spaced,
ordered, or $<\tau$). For arbitrary points $s^*=(x^*,t^*)\in\K$, the
belief over $q(s^*)$ is a Gaussian with mean function $\mu_\tau$, and
co-variance function $\Sigma_\tau$ \citep{RasmussenWilliams}
\begin{equation}
  \label{eq:2a}
  \begin{aligned}
    \mu_\tau(s^{*} _i) &= \vec{k} (s^* _i,\vec{S}_0) [K(\vec S_0,\vec
    S_0) + \sigma_q ^2\Id]^{-1} \vec{y}_0 \\
    \Sigma_\tau(s^* _i,s^* _j) &= k(s^* _i,s^* _j) - \vec{k}
    (s^* _i,\vec{S}_0) [K(\vec S_0,\vec S_0) + \sigma_q ^2 \Id]^{-1}
    \vec{k} (\vec{S}_0,s^* _j)
  \end{aligned}
\end{equation}
where $K(\vec S_0,\vec S_0)$ is the Gram matrix with elements $K_{ab}
= k(s_a,s_b)$. We will abbreviate $K_0\equiv [K(\vec S_0,\vec S_0) +
\sigma_y ^2 \Id]$ from here on. The co-vector $\vec{k}
(s^*,\vec{S}_0)$ has elements $\vec{k}_{i} = k(s^*,s_i)$ and will be
shortened to $\vec{k}_0$. How does this belief \emph{change} as time
moves from $\tau$ to $\tau+\d t$? If $\d t\to 0$, the chance of
acquiring a datapoint $y_\tau$ in this time is $\lambda \d
t$. Marginalising over this Poisson stochasticity, we expect one
sample with probability $\lambda \d t$, two samples with $(\lambda \d
t)^2$ and so on. So the mean after $\!\d t$ is expected to be
\begin{equation}
  \label{eq:7}
  \mu_{\tau+\d t} =
  \lambda \d t \left(
    \vec{k}_0, k_\tau \right) 
\begin{pmatrix}
  K_0 & \vec{\xi}_\tau \\ \vec{\xi}_\tau \Trans & \kappa_\tau
\end{pmatrix}^{-1}
\begin{pmatrix}
  \vec{y}_0\\ y_\tau \end{pmatrix} + (1-\lambda\d t - \O(
\lambda\d t)^2) \cdot
\vec{k} _0 K_0 ^{-1} \vec{y}_0 + \O(\lambda \d t)^2
\end{equation} 
where we have defined the map $k_{\tau} =k(s^*,s_\tau)$, the vector
$\vec{\xi}_{\tau}$ with elements $\xi_{\tau,i}=k(s_i,s_\tau)$, and the
scalar $\kappa_\tau=k(s_\tau,s_\tau) + \sigma_q ^2$. Algebraic
re-formulation yields \begin{equation}
  \label{eq:8}
  \mu_{\tau+\d t} = \vec{k}_0 K_0 ^{-1} \vec{y}_0 + \lambda
  (k_t - {\vec{k}_0}\Trans K_0 ^{-1}
  \vec{\xi}_t)(\kappa_t - \vec{\xi}_t \Trans K_0 ^{-1}\vec{\xi}_t)^{-1} (y_t -
  \vec{\xi}_t \Trans K_0 ^{-1} \vec{y}_0) \d t.
\end{equation}
Note that $\vec{\xi}_\tau \Trans K_0 ^{-1} \vec{y}_0=\mu(s_\tau)$, the
mean prediction at $s_{\tau}$ and $(\kappa_\tau - \vec{\xi}_\tau
\Trans K_0 ^{-1}\vec{\xi}_\tau)=\sigma^2 _q + \Sigma(s_\tau,s_\tau)$,
the marginal variance there. Hence, we can define scalars
$\bar{\Sigma},\bar{\sigma}$ and write
\begin{equation}
  \label{eq:6}
  \frac{(y_\tau - \vec{\xi}_\tau \Trans K_0 ^{-1}
    \vec{y}_0)}{(\kappa_\tau - \vec{\xi}_\tau \Trans K_0
    ^{-1}\vec{\xi}_\tau)^{1/2}} = \frac{[\Sigma^{1/2}\Omega](s_\tau) +
    \sigma \omega}{[\Sigma(s_\tau,s_\tau) + \sigma^2]^{1/2}}\equiv
  \bar{\Sigma}^{1/2} _\tau \Omega + \bar{\sigma}_\tau \omega
  \quad\text{with}\quad \omega \sim \N(0,1).
\end{equation}
So the change to the mean consists of a deterministic but uncertain
change whose effects accumulate linearly in time, and a
\emph{stochastic} change, caused by the independent noise process,
whose \emph{variance} accumulates linearly in time (in truth, these
two points are considerably subtler, a detailed proof is left out for lack of
space). We use the Wiener \citep{wiener1923differential} measure
$\d\omega$ to write
\begin{equation}
  \label{eq:5}
  \begin{aligned}
    \d \mu_{s_\tau}(s^*) &= \lambda\frac{k_\tau -
    {\vec{k}_0}\Trans K_0 ^{-1} \vec{\xi}_\tau}{(\kappa_\tau -
    \vec{\xi}_\tau \Trans K_0 ^{-1} \vec{\xi}_\tau)^{-1/2}} \frac{[\Sigma^{1/2}\Omega](s_\tau) +
    \sigma \omega}{[\Sigma(s_\tau,s_\tau) + \sigma^2]^{1/2}} \d t
    \equiv \lambda L_{s_\tau} (s^*) [\bar{\Sigma}^{1/2} _{\tau} \Omega
    \d t+
    \bar{\sigma}_\tau \d\omega]
  \end{aligned}
\end{equation}
where we have implicitly defined the \emph{innovation function}
$L$. Note that $L$ is a function of both $s^*$ and $s_\tau$. A similar
argument finds the change of the covariance function to be the
\emph{deterministic} rate
\begin{equation}
  \label{eq:10}
\begin{aligned}
  \d \Sigma_{s_\tau}(s^* _i,s^* _j) &= - \lambda L_{s_\tau}(s_i
  ^*)L\Trans _{s_\tau}(s_j ^*) \d t.
\end{aligned}
\end{equation}
So the dynamics of learning consist of a deterministic change to the
covariance, and both deterministic and stochastic changes to the mean,
both of which are samples a Gaussian processes with covariance
function proportional to $LL\Trans$. This separation is a fundamental
characteristic of GPs (it is the nonparametric version of a more
straightforward notion for finite-dimensional Gaussian beliefs, for
data with known noise magnitude). % $L\d \omega$ will be called the
% \emph{innovation process} in extension of the finite-dimensional
% concept \citep[e.g.][]{kailath1968innovations}.  Its significance here
% is that it allows a joint dynamic description of physical and learning
% system.

We introduce the belief-augmented space $\H$ containing states
$z(\tau)\equiv [x(\tau),\tau,\mu_q ^\tau (s),\mu_{f1} ^{\tau},\dots,
\mu_{fD} ^{\tau},$ $\Sigma_q ^\tau (s,s'),\Sigma_{f1}
^{\tau},\dots,\Sigma_{fD} ^{\tau}]$. Since the means and covariances
are functions, $\H$ is infinite-dimensional. Under our beliefs,
$z(\tau)$ obeys a stochastic differential equation of the form
\begin{equation}
  \label{eq:12}
  \d z = [\sA(z) + \sB(z)u + \sC(z) \Omega]\d t + \sD(z) \d\omega
\end{equation}
with \emph{free dynamics} $\sA$, \emph{controlled dynamics}
$\sB u$, \emph{uncertainty} operator $\sC$, and \emph{noise}
operator $\sD$
\begin{gather}
  \label{eq:13}
  \sA = \left[ \mu_f ^\tau (z_x,z_t)\;,\;1\;,\; 0\;,\;0\;,\;\dots\;,\;
    0\;,\;- \lambda L_qL_q\Trans \;,\; -\lambda L_{f1}L_{f1}\Trans
    \;,\; \dots
    \;,\; -\lambda L_{fD}L_{fD}\Trans \right]; \\
  \nonumber %\label{eq:13a}
  \sB = [g(s^*),0,0,0,\dots]; \quad \sC = \operatorname{diag}
  (\Sigma^{1/2} _{f\tau},0,\lambda L_q \bar{\Sigma}^{1/2}_q,\lambda
  L_{f1}\bar{\Sigma}^{1/2}_{f1},\dots,\lambda
  L_{fD}\bar{\Sigma}^{1/2}_{fd},0,\dots,0);\\ \sD =
  \operatorname{diag}(0,0,\lambda L_q\bar{\sigma}_q,\lambda
  L_{f1}\bar{\sigma}_{f1},\dots,\lambda
  L_{fD}\bar{\sigma}_{fD},0,\dots,0)
\end{gather}
The \emph{value} -- the expected cost to go -- of any state $s^*$ is
given by the Hamilton-Jacobi-Bellman equation, which follows from
Bellman's principle and a first-order expansion, using
Eq. \eqref{eq:3}:
\begin{multline}
  \label{eq:15}
  \raisetag{1.65cm}
%  \begin{aligned}
  v(z_\tau) = \min_u \left\{ \iint \left[\left(\mu_q(s_\tau) +
        \Sigma_{q\tau} ^{1/2}\Omega_q + \sigma_q\omega_q +
        \frac{1}{2}u\Trans R^{-1} u\right)\d t + v(z_{\tau+\d
        t})\right]
    \d\omega \d\Omega\right\}\\
  = \min_u \bigg\{ \int \mu_{q} ^\tau + \Sigma_{q\tau} ^{1/2}\Omega_q
  + \frac{1}{2}u\Trans R^{-1} u + \frac{v(z_\tau)}{\d t} + \frac{\de
    v}{\de t} +[\sA+\sB u + \sC\Omega] \Trans \nabla v
  + %\frac{1}{2} \tr[\sC\Trans (\nabla^2 v) \sC] +
  \frac{1}{2}\tr[\sD\Trans (\nabla^2 v) \sD] \d\Omega \bigg\}\d t % \\
  % \gamma^{-1} v(z) = \min_u \left\{\frac{1}{2}u\Trans R^{-1} u +
  %   u\Trans\sB(z)\Trans\nabla v\right\} + \mu_q ^\tau +
  % \sA(z)\Trans\nabla v
  % + %\frac{1}{2}\tr\left[\sC(z)\Trans (\nabla^2 v)\sC(z)\right] +
  % \frac{1}{2}\tr[\sD\Trans (\nabla^2 v) \sD]
% % \end{aligned}
\end{multline}
Integration over $\omega$ can be performed with ease, and removes the
\emph{stochasticity} from the problem; The \emph{uncertainty} over
$\Omega$ is a lot more challenging. Because the distribution over
future losses is correlated through space and time, $\nabla v$,
$\nabla^2 v$ are functions of $\Omega$, and the integral is nontrivial. But there
are some obvious approximate approaches. For example, if we
(inexactly) swap integration and minimisation, draw samples $\Omega^i$
and solve for the value for each sample, we get an ``average optimal
controller''. This over-estimates the actual sum of future rewards by
assuming the controller has access to the true system. It has the
potential advantage of considering the actual optimal controller for
every possible system, the disadvantage that the average of optima
need not be optimal for any actual solution. On the other hand, if we
ignore the correlation between $\Omega$ and $\nabla v$, we can
integrate \eqref{eq:18} locally, all terms in $\Omega$ drop out and we
are left with an ``optimal average controller'', which assumes that
the system locally follows its average (mean) dynamics. This cheaper
strategy was adopted in the following. Note that it is myopic, but not
greedy in a simplistic sense -- it does take the effect of learning
into account. It amounts to a ``global one-step look-ahead''. One
could imagine extensions that consider the influence of $\Omega$ on
$\nabla v$ to a higher order, but these will be left for future
work. Under this first-order approximation, analytic minimisation over
$u$ can be performed in closed form, and bears
\begin{equation}
  \label{eq:9}
  u(z) = - R \sB(z)\Trans \nabla v(z) = - R g(x,t)\Trans \nabla_x v(z).
\end{equation}
The \emph{optimal} Hamilton-Jacobi-Bellman equation is then
\begin{equation}
  \label{eq:17}
%   \gamma^{-1} v(z) = \mu_q ^\tau + \Sigma_{q\tau}^{1/2}\Omega_q + (\sA+\sC\Omega)\Trans \nabla v - \frac{1}{2} [\nabla v]\Trans \sB
%   R \sB\Trans \nabla v %+ \frac{1}{2} \tr \left[ \sC\Trans [\nabla^2
%                        %v] \sC\right] 
% + \frac{1}{2} \tr \left[ \sD\Trans (\nabla^2 v)
%     \sD\right].
  \gamma^{-1} v(z) = \mu_q ^\tau + \sA\Trans \nabla v - \frac{1}{2} [\nabla v]\Trans \sB
  R \sB\Trans \nabla v %+ \frac{1}{2} \tr \left[ \sC\Trans [\nabla^2
                       %v] \sC\right] 
+ \frac{1}{2} \tr \left[ \sD\Trans (\nabla^2 v)
    \sD\right].
\end{equation}
A more explicit form emerges upon re-inserting the definitions of
Eq. \eqref{eq:13} into Eq. \eqref{eq:17}:
\begin{multline}
  \label{eq:18}
  \gamma^{-1} v(z) = [\mu_q ^\tau %+ \Sigma_{q\tau}^{1/2}\Omega_q](z_x,z_t) \\
  + \underbrace{\big[ \mu_f ^\tau %+ \Sigma_{f\tau} ^{1/2}\Omega_f)
    (z_x,z_t)\nabla_x + \nabla_t \big]
    v(z)}_{\text{free drift cost}} 
  -\underbrace{\frac{1}{2} [\nabla_x
    v(z)]\Trans g\Trans(z_x,z_t) R g(z_x,z_t) \nabla_x
    v(z)}_{\text{control benefit}}
  \\  + \sum_{c=q,f_1,\dots,f_D} -\underbrace{\lambda\big[
    %L_c\Sigma_{c\tau} ^{1/2}\Omega_c \nabla_{\mu_c} -
    L_{c}L_{c}\Trans \nabla_{\Sigma_{c}} \big]v(z) }_{\text{exploration
    bonus}}
  + \underbrace{\frac{1}{2} \lambda^2
    \bar\sigma_c ^2 \big[ L_{fd} \Trans (\nabla^2 _{\mu_{fd}} v(z))
    L_{fd}\big] }_{\text{diffusion cost}}
\end{multline}
Equation \eqref{eq:18} is the central result: Given Gaussian priors on
nonlinear control-affine dynamic systems, up to a first order
approximation, optimal reinforcement learning is described by an
infinite-dimensional second-order partial differential equation. It
can be interpreted as follows (labels in the equation, note the
negative signs of ``beneficial'' terms): The value of a state
comprises the immediate utility rate; the effect of the free drift
through space-time and the benefit of optimal control; an
\emph{exploration bonus} of learning, and a \emph{diffusion cost}
engendered by the measurement noise. The first two lines of the right
hand side describe effects from the phase space-time subspace of the
augmented space, while the last line describes effects from the belief
part of the augmented space. The former will be called
\emph{exploitation terms}, the latter \emph{exploration terms}, for
the following reason: If the first two lines line dominate the right
hand side of Equation \eqref{eq:18} in absolute size, then future
losses are governed by the physical sub-space -- caused by exploiting
knowledge to control the physical system. On the other hand, if the
last line dominates the value function, exploration is more important
than exploitation -- the algorithm controls the physical space to
increase knowledge. To my knowledge, this is the first differential
statement about reinforcement learning's two objectives. Finally, note
the role of the sampling rate $\lambda$: If $\lambda$ is very low,
exploration is useless over the discount
horizon. % If it is very large, the system can be identified in a short
% time, then controlled optimally.

Even after these approximations, solving Equation \eqref{eq:18} for
$v$ remains nontrivial for two reasons: First, although the
vector product notation is pleasingly compact, the mean and covariance
functions are of course infinite-dimensional, and what looks like
straightforward inner vector products are in fact integrals. For
example, the average exploration bonus for the loss, writ large, reads
\begin{equation}
  \label{eq:19}
  -\lambda L_q L_q \Trans \nabla_{\Sigma_q} v(z) = -\iint_\K \lambda
  L^{(q)} _{s_\tau}(s^* _i) L^{(q)} _{s_\tau}(s^* _j) \frac{\de v(z)}{\de \Sigma(s^* _i,s^* _j)} \d s_i
  ^* \d s_j ^* .
\end{equation}
(note that this object remains a function of the state $s_\tau$). For
general kernels $k$, these integrals may only be solved
numerically. However, for at least one specific choice of kernel
(square-exponentials) and parametric Ansatz, the required integrals
can be solved in closed form. This analytic structure is so
interesting, and the square-exponential kernel so widely used that the
``numerical'' part of the paper (Section \ref{sec:numer-solv-hamilt})
will restrict the choice of kernel to this class.
% The necessary derivations are a marginal
% contribution of this paper. Technical and lengthy, they can be found
% in the supporting material.

The other problem, of course, is that Equation \eqref{eq:18} is a
nontrivial differential Equation. Section \ref{sec:numer-solv-hamilt}
presents \emph{one, initial} attempt at a numerical solution that
should not be mistaken for a definitive answer. Despite all this, Eq.\
\eqref{eq:18} arguably constitutes a useful gain for Bayesian
reinforcement learning: It replaces the intractable definition of the
value in terms of future trajectories with a differential
equation. This raises hope for new approaches to reinforcement learning,
based on numerical analysis rather than sampling.

\subsubsection*{Digression: Relaxing Some Assumptions}
\label{sec:relax-some-assumpt}

This paper only applies to the specific problem class of Section
\ref{sec:class-learn-probl}. Any generalisations and extensions are
future work, and I do not claim to solve them. But it is instructive
to consider some easier extensions, and some harder ones: For example,
it is intractable to simultaneously learn both $g$ and $f$
\emph{nonparametrically}, if only the actual transitions are observed,
because the beliefs over the two functions become infinitely dependent
when conditioned on data. But if the belief on either $g$ or $f$
is \emph{parametric} (e.g. a general linear model), a joint belief on
$g$ and $f$ is tractable \citep[see][\textsection
2.7]{RasmussenWilliams}, in fact straightforward. Both the quadratic
control cost $\propto u\Trans Ru$ and the control-affine form
($g(x,t)u$) are relaxable assumptions -- other parametric forms are
possible, as long as they allow for analytic optimization of
Eq.~\eqref{eq:15}. On the question of learning the kernels for
Gaussian process regression on $q$ and $f$ or $g$, it is clear that
standard ways of inferring kernels
\citep{RasmussenWilliams,murray2010slice} can be used without
complication, but that they are not covered by the notion of optimal
learning as addressed here.

\section{Numerically Solving the Hamilton-Jacobi-Bellman Equation}
\label{sec:numer-solv-hamilt}

Solving Equation \eqref{eq:17} is principally a problem of numerical
analysis, and a battery of numerical methods may be considered. 
This section reports on one specific Ansatz, a Galerkin-type
projection analogous to the one used in \citep{simpkins2008optimal}.
For this we break with the generality of previous sections and
assume that the kernels $k$ are given by square exponentials
$k(a,b)=k_\SE(a,b;\theta,S)=\theta^2\exp(-\frac{1}{2}(a-b)\Trans
S^{-1}(a-b))$ with parameters $\theta,S$. As discussed above, we
approximate by setting $\Omega=0$. We find an approximate
solution through a factorizing parametric Ansatz: Let the value of any
point $z\in\H$ in the belief space be given through a set of
parameters $\vec{w}$ and some nonlinear \emph{functionals}
$\vec{\phi}$, such that their contributions separate over phase space,
mean, and covariance functions:
\begin{equation}
  \label{eq:20}
  v(z) = \sum_{e=x,\Sigma_q,\mu_q,\Sigma_f,\mu_f} \vec{\phi}_e(z_e) \Trans \vec{w}_e \qquad
  \text{with } \vec{\phi}_e, \vec{w}_e \in \Re^{N_e}
\end{equation}
This projection is obviously restrictive, but it should be compared to
the use of radial basis functions for function approximation, a
similarly restrictive framework widely used in reinforcement learning.
The functionals $\phi$ have to be chosen conducive to the form of Eq.\
\eqref{eq:18}. For square exponential kernels, one convenient choice
is
\begin{align}
  \label{eq:21}
  \phi^a _s(z_s) &= k(s_z,s_a;\theta_a,S_a)\\
  \label{eq:21a}
  \phi^b _\Sigma(z_\Sigma) &= \iint_\K [\Sigma_z(s^* _i,s^* _j)-k(s_i
  ^*,s_j ^*)]
  k(s^* _i,s_b;\theta_b,S_b)   k(s^*
  _j,s_b;\theta_b,S_b)\d s^* _i \d s^* _j \quad \text{and}\\
  \label{eq:21b}
  %\phi^c _\mu(z_\mu) &= \int_\K \frac{1}{2}\mu^2 _z(s^*)
  %k(s^*,s_c,\theta_c,S_c)\d s^*
  \phi^c _\mu(z_\mu) &= \iint_\K \mu _z(s^* _i) \mu _z(s^* _j)
  k(s^* _i,s_c,\theta_c,S_c) k(s^* _j,s_c,\theta_c,S_c) \d s^* _i \d
  s^* _j 
\end{align}
(the subtracted term in the first integral serves only numerical
purposes). With this choice, the integrals of Equation \eqref{eq:18}
can be solved analytically (solutions left out due to space
constraints). %(see the
% supplementary material. , which also mentions another option,
% kernels on \emph{datasets} rather than the induced mean and
% covariance functions, in passing)
The approximate Ansatz turns Eq. \eqref{eq:18} into an algebraic
equation quadratic in $\vec{w}_x$, linear in all other $\vec{w}_e$:
\begin{equation}
  \label{eq:11}
  \frac{1}{2} \vec{w}_x\Trans \vec{\Psi}(z_x) \vec{w}_x - q(z_x) 
  + \sum_{e=x,\mu_q,\Sigma_q,\mu_f,\Sigma_f} \vec{\Xi}^e (z_e) \vec{w}_e = 0
\end{equation}
using co-vectors $\vec{\Xi}$ and a matrix $\vec{\Psi}$ with elements
%(Dirac's $\delta$ replaces the trace operation)
\begin{equation}
  \label{eq:14}
  \begin{aligned}
    \Xi^{x} _a(z_s) &= \gamma^{-1} \phi_s ^a(z_s) - f(z_x)\Trans
    \nabla_x \phi^a _s(z_s) -\nabla_t \phi^a _s(z_s)\\
    \Xi^\Sigma _a(z_\Sigma) &= \gamma^{-1} \phi_\Sigma ^a (z_\Sigma) +
    \lambda \iint_\K L_{s_\tau}(s^* _i) L_{s_\tau}(s^* _j) \frac{\de
      \phi_\Sigma(z_\Sigma)}{\de
      \Sigma_z(s^* _i, s^* _j)}\d s^* _i \d s^* _j\\
    \Xi^\mu _a (z_\mu) &= \gamma^{-1} \phi_\mu ^a(z_\mu) -
    \frac{\lambda^2\bar{\sigma}^2 _{s_\tau}}{2} \iint_\K %\delta(s^* _i - s^* _j)
    L_{s_\tau}(s^* _i) L_{s_\tau}(s^* _j) \frac{\de^2 \phi_\mu
      ^a(z_\mu)}{\de \mu_z(s^* _i)\de \mu_z(s^* _j)}
    \d s^* _i \d s^* _j \\
    \Psi(z) _{k\ell} &=[\nabla_x \phi^k _s(z)]\Trans g(z_x) R
    g(z_x)\Trans [\nabla_x \phi_s ^\ell(z)]
  \end{aligned}
\end{equation}
Note that $\Xi_{\mu}$ and $\Xi_{\Sigma}$ are both functions of the
physical state, through $s_\tau$. It is through this functional
dependency that the value of information is associated with the
physical phase space-time. To solve for $\vec{w}$, we simply choose a
number of evaluation points $\vec{z}_\eval$ sufficient to
constrain the resulting system of quadratic equations, and then find
the least-squares solution $\vec{w}_\text{opt}$ by function
minimisation, using standard methods, such as Levenberg-Marquardt
\citep{marquardt1963algorithm}. A disadvantage of this approach is
that is has a number of degrees of freedom $\vec{\Theta}$, such as the
kernel parameters, and the number and locations $\vec{x}_a$ of the
feature functionals. Our experiments (Section \ref{sec:experiments})
suggest that it is nevertheless possible to get interesting results
simply by choosing these parameters heuristically.
% \begin{algorithm}
% \label{alg:1}
% \newcommand{\mycft}[1]{{\footnotesize  #1}}
% \SetCommentSty{mycft}
% %  \DontPrintSemicolon
%   \SetKwFunction{GenerateBasePoints}{GenerateNumericalBasis}
%   \SetKwFunction{EvaluateDynamics}{GPRegression}
%   \SetKwFunction{ConstructLinearMaps}{ConstructLinearMaps}
%   \SetKwFunction{Minimize}{Minimize}
%   \caption{Approximate Bayes-Optimal Learning Controller for Gaussian Systems}
%   \KwData{Observations $S_q$,$Y_q$,$S_f$,$Y_f$ for utilities and
%     dynamics. Scales $\gamma, R, k_q, \vec{k}_f, \vec{h}_v$, Noises
%     $\xi_q,\vec{\xi}_f$} \KwResult{Optimal Control $u=-R[g(s_\tau)]\Trans
%     \nabla_x \vec{\phi}_x(s_\tau)\vec{w}_x$}
%   \Begin{
%     $[\vec{\Theta},\vec{z}_\eval] \gets$
%     \GenerateBasePoints{}  \tcp*{Heuristic. May be cached and reused}
%     $[L_q,L_f,\hat q,\hat f]\gets$ \EvaluateDynamics{$S_q,Y_q,S_f,Y_f$}
%     \tcp*{Eq.\ \eqref{eq:13}}
%     $[\vec{\Xi},\vec{\Psi}]\gets$
%     \ConstructLinearMaps{\rm $L_q,L_f,\hat q,\hat f,\vec{\Theta},\vec{z}_\eval$} 
%     \tcp*{Eq.\ \eqref{eq:14}, Suppl. Mat.}
%     $\vec{w}_\text{opt} \gets$ \Minimize{\rm $\|\vec{\Xi}\vec{w} - r(\vec{z}_\eval) -
%       \frac{1}{2}\vec{w}\Trans \Psi \vec{w}\|^2$}
%     \tcp*{standard problem}
%   }
% \end{algorithm}
% \vspace{-4mm} 

\section{Experiments}
\label{sec:experiments}

% The first experiment uses a simple, one-dimensional environment, to
% demonstrate some aspects that are perhaps not obvious. It is followed
% by a sample application, comparing to other algorithms.

\subsection{Illustrative Experiment on an Artificial Environment}
\label{sec:illustr-exampl-an}

\begin{figure}[ht]
%  \centering
% This file was created by matlab2tikz v0.1.0.
% Copyright (c) 2008--2011, Nico Schlömer <nico.schloemer@gmail.com>
% All rights reserved.
% 
% The latest updates can be retrieved from
%   http://www.mathworks.com/matlabcentral/fileexchange/22022-matlab2tikz
% where you can also make suggestions and rate matlab2tikz.
% 
\begin{tikzpicture}

\begin{axis}[%
name=plot1,
scale only axis,
width=120pt, %14.7455in,
height=100pt, %14.7455in,
xmin=-1, xmax=101,
ymin=-51, ymax=51,
ylabel={$x$},
axis on top]
\addplot graphics [xmin=-1, xmax=101, ymin=-51, ymax=51] {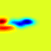};
\addplot [
color=black,
only marks,
mark=*,
mark options={solid}
]
coordinates{ (0,0) (0.5,0.0771771) (1,0.104918) (1.5,0.29822) (2,0.624107) (2.5,1.02507) (3,1.01625) (3.5,0.562104) (4,0.468946) (4.5,0.533325) (5,1.51685) (5.5,2.38482) (6,3.13762) (6.5,3.27053) (7,3.01168) (7.5,2.92546) (8,2.75101) (8.5,2.53066) (9,2.48974) (9.5,2.19136) (10,2.04458) (10.5,1.46624) (11,1.39722) (11.5,1.52597) (12,1.63814) (12.5,1.68437) (13,2.09216) (13.5,2.27412) (14,2.44226) (14.5,2.79805) (15,2.93694) (15.5,2.8669) (16,2.72378) (16.5,2.75475) (17,3.03758) (17.5,3.16912) (18,3.41737) (18.5,3.4545) (19,3.37621) (19.5,3.29812) (20,3.31329) (20.5,3.44117) (21,3.32786) (21.5,3.41737) (22,3.54102) (22.5,3.6878) (23,3.77748) (23.5,3.8463) (24,3.88693) (24.5,4.02) (25,4.25419) (25.5,4.38988) (26,4.61014) (26.5,4.41896) (27,4.36657) (27.5,4.21233) (28,4.1827) (28.5,4.21284) (29,4.11034) (29.5,4.22165) (30,4.41712) (30.5,4.38795) (31,4.30483) (31.5,4.28814) (32,4.19856) (32.5,4.3457) (33,4.47256) (33.5,4.74666) (34,4.85339) (34.5,4.76011) (35,4.78882) (35.5,4.74131) (36,4.67121) (36.5,4.54959) (37,4.64145) (37.5,4.58855) (38,4.42731) (38.5,4.40486) (39,4.42015) (39.5,4.3387) (40,4.40519) (40.5,4.39555) (41,4.49767) (41.5,4.51595) (42,4.50398) (42.5,4.45664) (43,4.6127) (43.5,4.89197) (44,5.14561) (44.5,5.48714) (45,5.70061) (45.5,5.78541) (46,5.88997) (46.5,6.07012) (47,6.15663) (47.5,6.37308) (48,6.61877) (48.5,6.88044) (49,7.24586) (49.5,7.7008) (50,7.86549)
};

\addplot [
color=green,
only marks,
mark=diamond*,
mark options={solid,fill=black}
]
coordinates{ (50.5,8.00171)
};

\end{axis}

% colorbar
\begin{axis}[%
axis on top,
at=(plot1.right of south east), anchor=left of south west,
width=8pt, height=100pt,
scale only axis,
xmin=0, xmax=1,
ymin=-1.20432, ymax=0.834301,
xtick=\empty, yticklabel pos=right,
xticklabels={\empty}]
\addplot graphics [xmin=0, xmax=1, ymin=-1.204315e+000, ymax=8.343015e-001] {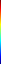};
\end{axis}
\end{tikzpicture}
% This file was created by matlab2tikz v0.1.0.
% Copyright (c) 2008--2011, Nico Schlömer <nico.schloemer@gmail.com>
% All rights reserved.
% 
% The latest updates can be retrieved from
%   http://www.mathworks.com/matlabcentral/fileexchange/22022-matlab2tikz
% where you can also make suggestions and rate matlab2tikz.
% 
\begin{tikzpicture}

\begin{axis}[%
name=plot1,
scale only axis,
width=120pt,
height=100pt,
xmin=-1, xmax=101,
ymin=-51, ymax=51,
axis on top]
\addplot graphics [xmin=-1, xmax=101, ymin=-51, ymax=51] {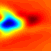};
\addplot [
color=green,
only marks,
mark=diamond*,
mark options={solid,fill=black}
]
coordinates{ (50.5,8.00171)
};

\end{axis}

% colorbar
\begin{axis}[%
axis on top,
at=(plot1.right of south east), anchor=left of south west,
width=8pt, height=100pt,
scale only axis,
xmin=0, xmax=1,
ymin=-8.62208, ymax=0.11029,
xtick=\empty, yticklabel pos=right,
xticklabels={\empty}]
\addplot graphics [xmin=0, xmax=1, ymin=-8.622084e+000, ymax=1.102904e-001] {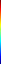};
\end{axis}
\end{tikzpicture}
% This file was created by matlab2tikz v0.1.0.
% Copyright (c) 2008--2011, Nico Schlömer <nico.schloemer@gmail.com>
% All rights reserved.
% 
% The latest updates can be retrieved from
%   http://www.mathworks.com/matlabcentral/fileexchange/22022-matlab2tikz
% where you can also make suggestions and rate matlab2tikz.
% 
\begin{tikzpicture}

\begin{axis}[%
name=plot1,
scale only axis,
width=120pt,
height=100pt,
xmin=-1, xmax=101,
ymin=-51, ymax=51,
ylabel={$x$},
xlabel={$t$},
axis on top]
\addplot graphics [xmin=-1, xmax=101, ymin=-51, ymax=51] {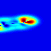};
\addplot [
color=green,
only marks,
mark=diamond*,
mark options={solid,fill=black}
]
coordinates{ (50.5,8.00171)
};

\end{axis}

% colorbar
\begin{axis}[%
axis on top,
at=(plot1.right of south east), anchor=left of south west,
width=8pt, height=100pt,
scale only axis,
xmin=0, xmax=1,
ymin=-0.121611, ymax=0.930507,
xtick=\empty, yticklabel pos=right,
xticklabels={\empty}]
\addplot graphics [xmin=0, xmax=1, ymin=-1.216111e-001, ymax=9.305072e-001] {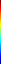};
\end{axis}
\end{tikzpicture}\hspace{4.75mm}
% This file was created by matlab2tikz v0.1.0.
% Copyright (c) 2008--2011, Nico Schlömer <nico.schloemer@gmail.com>
% All rights reserved.
% 
% The latest updates can be retrieved from
%   http://www.mathworks.com/matlabcentral/fileexchange/22022-matlab2tikz
% where you can also make suggestions and rate matlab2tikz.
% 
\begin{tikzpicture}

\begin{axis}[%
name=plot1,
scale only axis,
width=120pt,
height=100pt,
xmin=-1, xmax=101,
ymin=-51, ymax=51,
xlabel={$t$},
axis on top]
\addplot graphics [xmin=-1, xmax=101, ymin=-51, ymax=51] {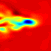};
\addplot [
color=green,
only marks,
mark=diamond*,
mark options={solid,fill=black}
]
coordinates{ (50.5,8.00171)
};

\end{axis}

% colorbar
\begin{axis}[%
axis on top,
at=(plot1.right of south east), anchor=left of south west,
width=8pt, height=100pt,
scale only axis,
xmin=0, xmax=1,
ymin=-1.02559, ymax=0.11551,
xtick=\empty, yticklabel pos=right,
xticklabels={\empty}]
\addplot graphics [xmin=0, xmax=1, ymin=-1.025588e+000, ymax=1.155102e-001] {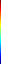};
\end{axis}
\end{tikzpicture}
\caption{State after 50 time steps, plotted over phase
  space-time. {\bfseries top left:} $\mu_q$ (blue is good). The belief
  over $f$ is not shown, but has similar structure.  {\bfseries top
    right:} value estimate $v$ at current belief: compare to next two
  panels to note that the approximation is relatively
  coarse. {\bfseries bottom left:} exploration terms. {\bfseries
    bottom right:} exploitation terms. At its current state (black
  diamond), the system is in the process of switching from
  exploitation to exploration (blue region in bottom right panel is
  roughly cancelled by red, forward cone in bottom left one).}
  \label{fig:colourful}
\end{figure}

As a simple example system with a one-dimensional state space, $f,q$
were sampled from the model described in Section
\ref{sec:class-learn-probl}, and $g$ set to the unit function. The
state space was tiled regularly, in a bounded region, with $231$
square exponential (``radial'') basis functions (Equation
\ref{eq:21}), initially all with weight $w_x ^i =0$. For the
information terms, only a single basis function was used for each term
(i.e.\ one single $\phi_{\Sigma q}$, one single $\phi_{\mu q}$, and
equally for $f$, all with very large length scales $S$, covering the
entire region of interest). As pointed out above, this does not imply
a trivial structure for these terms, because of the functional
dependency on $L_{s_\tau}$. Five times the number of parameters,
i.e. $N_\eval=1175$ evaluation points $z_\eval$ were sampled, at each
time step, uniformly over the same region. It is not intuitively clear
whether each $z_e$ should have its own belief (i.e.\ whether the
points must cover the belief space as well as the phase space), but
anecdotal evidence from the experiments suggests that it suffices to
use the current beliefs for all evaluation points. A more
comprehensive evaluation of such aspects will be the subject of a
future paper. The discount factor was set to $\gamma=50\mathrm{s}$,
the sampling rate at $\lambda=2/\mathrm{s}$, the control cost at
$10\mathrm{m^2}/(\$\mathrm{s})$. Value and optimal control were
evaluated at time steps of $\delta t=1/\lambda=0.5\mathrm{s}$.

Figure \ref{fig:colourful} shows the situation $50\mathrm{s}$ after
initialisation. % (The supplements contain a more revealing video of a
% second initialisation, see detailed comments in the appendix). 
The most noteworthy aspect is the nontrivial structure of exploration
and exploitation terms. Despite the simplistic parameterisation of the
corresponding functionals, their functional dependence on $s_\tau$
induces a complex shape. The system constantly balances exploration
and exploitation, and the optimal balance depends nontrivially on
location, time, and the actual value (as opposed to only uncertainty)
of accumulated knowledge. This is an important insight that casts
doubt on the usefulness of simple, local exploration boni, used in
many reinforcement learning algorithms.

Secondly, note that the system's trajectory does not necessarily
follow what would be the optimal path under full information. The
value estimate reflects this, by assigning low (good) value to regions
\emph{behind} the system's trajectory. This amounts to a sense of
``remorse'': If the learner would have known about these regions
earlier, it would have strived to reach them. But this is not a sign
of sub-optimality: Remember that the value is defined on the augmented
space. The plots in Figure \ref{fig:colourful} are merely a slice
through that space at some level set in the belief space. % The video in
% the supplementary material can be construed as a flight through this
% space along one learning trajectory.

\subsection{Comparative Experiment -- The Furuta Pendulum}
\label{sec:an-application}

\begin{figure}[ht]
\hfill
 \begin{tikzpicture}
   % base
    \draw (-1.5,0) to [controls=+(90:0.5) and +(90:0.5)] (1.5,0);
    \draw (-1.5,0) .. controls +(-90:0.5) and +(-90:0.5) .. (1.5,0);
    \draw (-1.5,0) .. controls +(-90:0.5) and +(-90:0.5) .. (1.5,0)
        -- (1.5,-1.5) .. controls +(-90:0.5) and +(-90:0.5) .. (-1.5,-1.5)
        -- (-1.5,0);

    % centre of rotation
    \node[circle,fill=black,minimum width=0.1cm,inner sep=0] at (0,0)
    {};
    % arm     
   \draw [|<->|] (0.11,-0.15) -- +(30:-1) node[midway,fill=white,inner
    sep=0pt] {\footnotesize $\ell_1$};
    \draw[ultra thick] (0,0) -- +(30:-1);
   \draw[dashed] (0,0) -- +(0:-1.8);
   \node[fill=white,inner sep=0pt] at (-0.65,-0.1) {\footnotesize
     $\theta_1$};

   \draw[->] (0,0) -- +(0,1) node[midway,right] {$u$};
    % pendulum
   \node[circle,fill=black,minimum width=0.1cm,inner sep=0] at (-0.86,-0.5) {};
   \draw[|<->|] (-0.2,0.05) ++ (30:-1) -- +(70:-2)
   node[midway,fill=white,inner sep=0pt] {\footnotesize $\ell_2$};
    \draw[ultra thick] (0,0) ++(30:-1) -- +(70:-2);
   \draw[dashed] (0,0) ++(30:-1) -- +(90:-1.5);
   \node[fill=white,inner sep=0pt] at (-1.1,-1.8) {\footnotesize $\theta_2$};
  \end{tikzpicture}
\hfill
  \begin{tabular}[b]{l r @{.} l @{$\pm$} l}
    \toprule
    Method & \multicolumn{3}{c}{cumulative loss} \\
    \midrule
    Full Information (baseline) & $4$&$4 $&$ 0.3$\\
    TD($\lambda$) & $6$&$401 $&$ 0.001$\\
    Kalman filter Optimal Learner & $6$&$408 $&$ 0.001$ \\
    %{\bfseries Gaussian process optimal learner} & $5$&$1 $&$ 0.3$ \\
    {\bfseries Gaussian process optimal learner} & $4$&$6 $&$ 1.4$ \\
    \bottomrule
    \vspace{3mm}
  \end{tabular}
\hfill
  \caption{The Furuta pendulum system: A pendulum of length $\ell_2$
    is attached to a rotatable arm of length $\ell_1$. The control
    input is the torque applied to the arm. {\bfseries Right:}
  cost to go achieved by different methods. Lower is better. Error
  measures are one standard deviation over five experiments.}
  \label{fig:furuta}
\end{figure}

The cart-and-pole system is an under-actuated problem widely studied
in reinforcement learning. For variation, this experiment uses a
cylindrical version, the pendulum on the rotating arm \citep{Furuta}.
The task is to swing up the pendulum from the lower resting point. The
table in Figure \ref{fig:furuta} compares the average loss of a
controller with access to the true $f,g,q$, but otherwise using
Algorithm 1, to that of an $\epsilon$-greedy TD$(\lambda)$ learner
with linear function approximation, Simpkins' et al.'s
\citep{simpkins2008optimal} Kalman method and the Gaussian process
learning controller (Fig. \ref{fig:furuta}). The linear function
approximation of TD$(\lambda)$ used the same radial basis functions as
the three other methods. None of these methods is free of assumptions:
Note that the sampling frequency influences TD in nontrivial ways
rarely studied (for example through the coarseness of the
$\epsilon$-greedy policy). The parameters were set to
$\gamma=5\mathrm{s}$, $\lambda=50/\mathrm{s}$. Note that reinforcement
learning experiments often quote total accumulated loss, which differs
from the discounted task posed to the learner. Figure \ref{fig:furuta}
reports actual discounted losses.  The GP method clearly outperforms
the other two learners, which barely explore. Interestingly, none of
the tested methods, not even the informed controller, achieve a stable
controlled balance, although the GP learner does swing up the
pendulum. This is due to the random, non-optimal location of basis
functions, which means resolution is not necessarily available where
it is needed (in regions of high curvature of the value function), and
demonstrates a need for better solution methods for Eq. \eqref{eq:18}.

There is of course a large number of other algorithms methods to
potentially compare to, and these results are anything but
exhaustive. They should not be misunderstood as a critique of any
other method. But they highlight the need for units of measure on
every quantity, and show how hard optimal exploration and exploitation
truly is. Note that, for time-varying or discounted problems, there is
no ``conservative'' option that cold be adopted in place of the
Bayesian answer.

\section{Conclusion}
\label{sec:conclusion}

Gaussian process priors provide a nontrivial class of reinforcement
learning problems for which optimal reinforcement learning reduces to
solving differential equations. Of course, this fact alone does not
make the problem easier, as solving nonlinear differential equations
is in general intractable. However, the ubiquity of differential
descriptions in other fields raises hope that this insight opens new
approaches to reinforcement learning. For intuition on how such
solutions might work, one specific approximation was presented, using
functionals to reduce the problem to finite least-squares parameter
estimation.

The critical reader will have noted how central the prior is for the
arguments in Section \ref{sec:optim-contr-learn}: The dynamics of the
learning process are predictions of future data, thus inherently
determined exclusively by prior assumptions. One may find this
unappealing, but there is no escape from it. Minimizing future loss
requires predicting future loss, and predictions are always in danger
of falling victim to incorrect assumptions. A finite initial
identification phase may mitigate this problem by replacing prior with
posterior uncertainty -- but even then, predictions and decisions will
depend on the model.

% The class of systems for which our method applies is not arbitrarily
% general, but provides reasonable descriptions for a wide range of
% physical systems. It even extends on some assumptions of classic
% reinforcement learning, e.g.\ by allowing dynamics and reward
% expectations to change over time. There are two arguments for
% ``structured'', probabilistic, approaches like the one presented
% here. The first is the immense complexity of the general reinforcement
% learning problem, which makes it unlikely that efficient --- in the
% probabilistic sense --- yet universal reinforcement learning
% algorithms for realistic systems will be found in the foreseeable
% future. The other is the impressive collection of successes that
% control theory has historically had using structured models.

% Our dynamics model may look restrictive from a learning
% theoretic perspective; but, as a learning algorithm, it is still
% fundamentally more flexible than a classic optimal control method.

The results of this paper raise new questions, theoretical and
applied. The most pressing questions concern better solution methods
for Eq. \eqref{eq:15}, in particular better means for taking the
expectation over the uncertain dynamics to more than first
order. There are also obvious probabilistic issues: Are there other
classes of priors that allow similar treatments? (Note some conceptual
similarities between this work and the BEETLE algorithm
\citep{poupart2006analytic}). To what extent can approximate inference
methods -- widely studied in combination with Gaussian process
regression -- be used to broaden the utility of these results?

% From the theoretical perspective, an intriguing query is how
% computational complexity interacts with prior assumptions: What is the
% class of reinforcement learning problems that can be solved in
% polynomial time? Another concern is the utility of approximate
% inference methods for expanding this class.

% On the theoretical side, a pressing but
% very challenging issue is wether any statements about stability are
% possible. Nonlinear systems show notoriously complex behaviour
% \citep{Slotine}, and the very generality of the Gaussian Process
% regression framework that is a virtue of our approach makes it hard to
% make any analytical statements about the stability of an arbitrary
% system fitting the framework.

\section*{Acknowledgments}
The author wishes to express his gratitude to Carl Rasmussen, Jan Peters, Zoubin
Ghahramani, Peter Dayan, and an anonymous reviewer, whose thoughtful comments
uncovered several errors and crucially improved this paper.

\newpage
\small{
\bibliographystyle{unsrt} %plainnat
\bibliography{../bibfile}
}

\end{document}